\pdfoutput=1

\documentclass[11pt]{article}

\usepackage[final]{acl}

\usepackage{times}
\usepackage{latexsym}
\usepackage{amsmath}

\usepackage[T1]{fontenc}

\usepackage[utf8]{inputenc}

\usepackage{microtype}

\usepackage{inconsolata}

\usepackage{graphicx}

\usepackage{booktabs}
\usepackage{listings}
\usepackage{xspace}
\usepackage{subcaption}
\usepackage{enumitem}
\usepackage{array}

\newif\ifshowcomments
\showcommentsfalse

\ifshowcomments
  \newcommand{\chenhao}[1]{\textcolor{blue}{[#1 ---\textsc{ct}]}}
  \newcommand{\karen}[1]{\textcolor{magenta}{[#1 ---\textsc{kz}]}}
  \newcommand{\todo}[1]{\textcolor{red}{[\textsc{todo} --- #1]}}
  \newcommand{\davis}[1]{\textcolor{green}{[#1 ---\textsc{dl}]}}
\else
  \newcommand{\chenhao}[1]{}
  \newcommand{\karen}[1]{}
  \newcommand{\todo}[1]{}
  \newcommand{\davis}[1]{}
\fi

\newcommand{\figref}[1]{Fig.~\ref{#1}}
\newcommand{\tbref}[1]{Table~\ref{#1}}
\newcommand{\secref}[1]{\S\ref{#1}}
\newcommand{\para}[1]{\vspace{0.5em}\noindent{\bf #1}}

\title{From Feedback to Checklists: Grounded Evaluation of AI-Generated Clinical Notes}

\author{\textbf{Karen Zhou\textsuperscript{1}\thanks{Work done during internship at Abridge.}}, 
  \textbf{John Giorgi\textsuperscript{2}},
  \textbf{Pranav Mani\textsuperscript{2}},
  \textbf{Peng Xu\textsuperscript{2}},\\
  \textbf{Davis Liang\textsuperscript{2}}, 
  \textbf{Chenhao Tan\textsuperscript{1}} 
  \\
  \textsuperscript{1}University of Chicago,
  \textsuperscript{2}Abridge
  \\
  \texttt{\{karenzhou, chenhao\}@uchicago.edu}\\
  \texttt{\{john, pranav, peng.xu, davis\}@abridge.com}
  }

\begin{document}
\maketitle
\begin{abstract}
AI-generated clinical notes are increasingly used in healthcare, but evaluating their quality remains a challenge due to high subjectivity and limited scalability of expert review. Existing automated metrics often fail to align with real-world physician preferences. 
To address this, we propose a pipeline that systematically distills real user feedback into structured checklists for note evaluation. 
These checklists are designed to be interpretable, grounded in human feedback, and enforceable by LLM-based evaluators.
Using deidentified data from over 21,000 clinical encounters (prepared in accordance with the HIPAA safe harbor standard) from a deployed AI medical scribe system, we show that our feedback-derived checklist outperforms a baseline approach in our offline evaluations in coverage, diversity, and predictive power for human ratings.
Extensive experiments confirm the checklist's robustness to quality-degrading perturbations, significant alignment with clinician preferences, and practical value as an evaluation methodology.
In offline research settings, our checklist offers a practical tool for flagging notes that may fall short of our defined quality standards.

\end{abstract}

\section{Introduction}
\label{sec:intro}

\begin{figure}[h]
    \centering
    \includegraphics[width=0.95\linewidth]{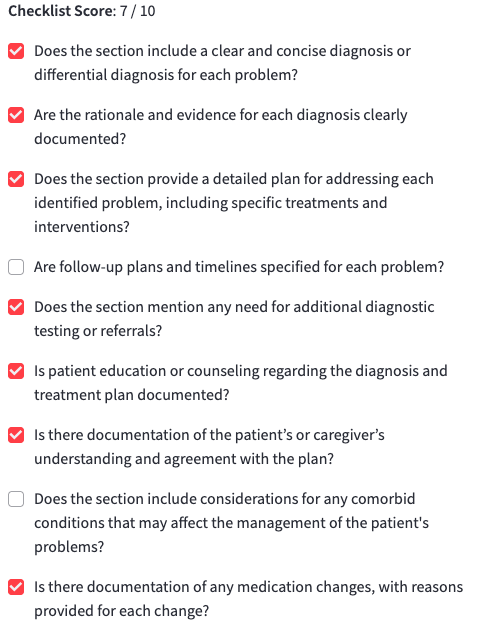}
    \caption{Example checklist questions for the Assessment and Plan section of a clinical note. The checklist score consists of the proportion of satisfied questions.
    }
    \label{fig:example}
\end{figure}

Evaluating the quality of AI-generated text is challenging, especially in domains where the text is highly specialized and requires expert knowledge. This is particularly true in the medical domain, specifically for AI-generated clinical notes. 
Automated metrics are scalable but often misaligned with clinical needs, relying on limited reference notes and penalizing harmless stylistic differences. Reference-free methods tend to focus narrowly on factuality \citep{xie-etal-2024-doclens}.
Meanwhile, human evaluation by clinical experts is generally high-quality but costly, inconsistent, and difficult to scale due to subjective preferences and varying documentation standards across specialties \citep{wang2025adaptingopensourcelargelanguage, Hanson2012QualityOO}. The diversity of clinical practice  makes defining a single set of evaluation criteria (e.g., a rubric) that applies across all notes difficult \citep{aci-bench, wang2024perspectiveadaptinggeneralistai, Croxford2025CurrentAF}.

Checklists are commonly employed in aviation and healthcare, used to 
improve safety and quality \citep{Reijers2017TowardsAS, Chaparro2019ChecklistsAR, 10.1145/308769.308798, doi:10.1177/001872089303500209, doi:10.1056/NEJMsa0810119} and to guide human- or LLM-evaluators in complex tasks.

To address the challenge of evaluating AI-generated clinical notes, we propose leveraging real user feedback to generate grounded checklists.
For each encounter, clinicians may provide free-form written feedback for the corresponding note.
Such user feedback reflects real issues clinicians face with AI-generated notes, providing valuable insights into the qualities that make a clinical note effective or lacking. By analyzing this feedback, we can automatically uncover the attributes associated with highly-rated notes.
These qualities can be compiled into structured checklists to effectively evaluate clinical notes. 
The checklist can be used 
with LLMs-as-a-Judge \citep{zheng2023judging},
to automatically flag notes that are lacking, at scale for checklists that are too long for humans to evaluate, or to support human evaluation by providing a rubric for the human evaluator to follow.
Thus, we tackle the following research question: can user feedback be systematically distilled into a note quality checklist?

To achieve these goals, we propose a method to automatically generate a checklist, given user feedback. The pipeline involves leveraging user feedback as context, prompting an LLM for candidate checklist questions, and then applying a series of refinement steps to select for the most salient questions. 
Finally, we conduct automatic evaluation of the final checklists, with plans for human validation and checking for generalization\footnote{Our checklists are not intended for clinical deployment or for altering patient care without clinician review.}. 
An example of how a checklist would be used for scoring a note can be seen in \figref{fig:example}. 

Our contributions are as follows:
\begin{itemize}[leftmargin=*, itemsep=0em]
    \item We propose a systematic pipeline for generating and refining checklists from real user feedback, with clear evaluation metrics.
    \item In our internal offline evaluation, the checklist appears more comprehensive, diverse, and predictive of human ratings than a baseline (zero-shot)
    checklist.
    \item We demonstrate the checklist's robustness to various quality-degrading perturbations (e.g., missing information, poor writing flow and organization, and redundancy and hallucinations)
    as well as significant alignment with expert preferences.
\end{itemize}

\section{Related Work}
\label{sec:related_work}

Additional work on checklist generation and evaluation are discussed in Appendix \secref{asec:related_work}.

\paragraph{Clinical note evaluation}
Guideline or rubric-based evaluation of clinical notes has been explored in several works \citep{yim-etal-2019-automatic}.
\citet{stetson2012assessing} is a 9-item rubric for medical documentation quality, a shorter version of the Physician Documentation Quality Instrument (PDQI) \citep{STETSON2008534}. It is scored on a 5-point Likert scale and has been used to evaluate AI-generated clinical notes 
\citep{amenyo2025assessmentaigeneratedpediatricrehabilitation, croxford2025developmentvalidationproviderdocumentation}. 
\citet{Eng2024APD} creates a high-level guideline for medical documentation best practices.
\citet{10.1136/amiajnl-2013-002321} is another physician-validated medical note rubric. These rubrics often have Likert scales as answer options, which introduces additional complexity  and ambiguity for human and LLM raters. 
Rubrics are also difficult to develop and score, due to subjectivity and variability of practice \citep{Croxford2025CurrentAF, oleson2024deepscorecomprehensiveapproachmeasuring}.
While these existing rubrics are static, we introduce an approach that is comprehensive of large-scale user concerns and dynamic to new feedback.

Other note evaluation works include training reward models \citep{wang2025processsupervisedrewardmodelsverifying}, 
or proposing reference-free evaluation on pre-defined desired attributes (typically factuality-based) \citep{kanithi2024mediccomprehensiveframeworkevaluating, xie-etal-2024-doclens}.
We are the first to propose a checklist-based evaluation approach grounded in real human feedback.

\paragraph{Checklist-based evaluation with LLM-as-a-Judge}
LLMs are increasingly used as automatic evaluators, due to their ability to handle nuance and scale. 
In particular, LLM-evaluators (or LLMs-as-a-Judge) are used to evaluate the quality of another LLM's generations.
However, LLM-evaluators have been shown to contain biases like increased position bias with more answer candidates
\citep{ye2024justice}, leading us to choose a binary checklist-based evaluation approach.

Works like \citet{goldberg2024usefulness} use LLMs to evaluate compliance with a pre-existing checklist, and many others have explored checklist-based evaluation with LLM-as-a-Judge for a variety of tasks \citep{lin2024wildbenchbenchmarkingllmschallenging, que2024hellobenchevaluatinglongtext, chu2025thinkworkbettercombining}. It has been extensively demonstrated that checklists can be used to improve the reliability of LLM-evaluators and agreement with human judgments \citep{lee2025checkevalreliablellmasajudgeframework,
pereira2024checkevalchecklistbasedapproachevaluating, 
cook2024tickingboxesgeneratedchecklists, 
wei2025rocketeval, 
li2025hypoevalhypothesisguidedevaluationnatural}.

\begin{figure*}
    \centering
    \includegraphics[width=0.95\linewidth]{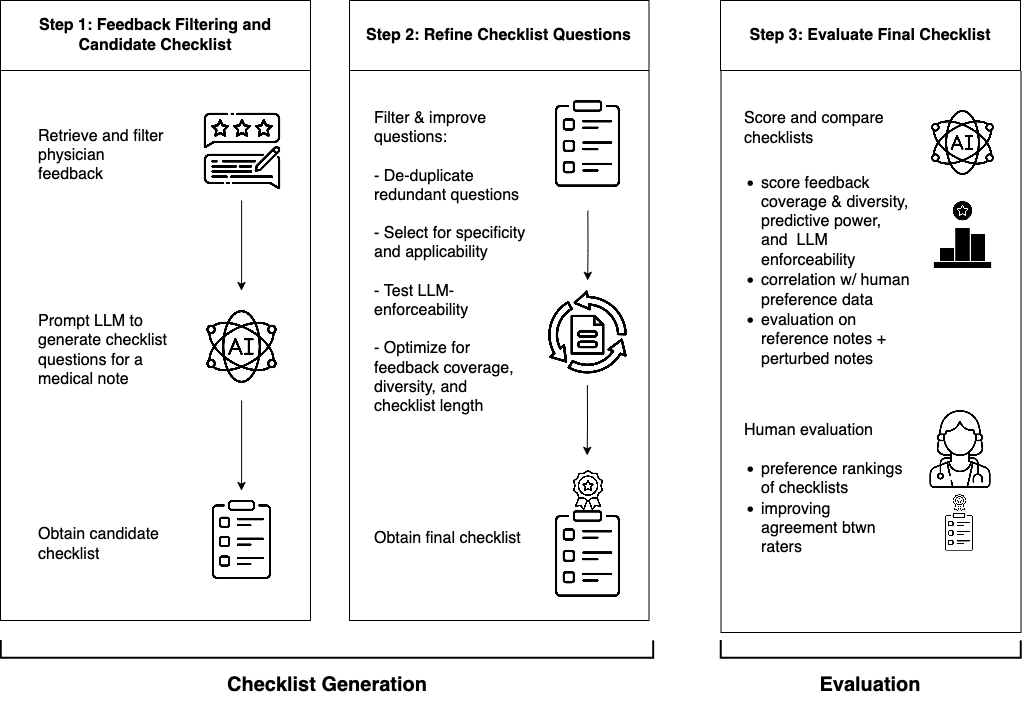}
    \caption{Proposed end-to-end pipeline.}
    \label{fig:pipeline}
\end{figure*}

\section{Data}
\label{sec:data}

We leverage a mix of proprietary data (user feedback and note ratings) from a commercial Ambient AI company and open-source datasets (clinical notes). We use: 1) approximately 22,000 real-world clinical encounters containing free-text user feedback and 1–5 star ratings
(see \tbref{tab:feedback} for synthetic examples), 2) 506 pairs of clinical notes rated by clinical experts in pairwise comparisons, and 3) 207 reference clinical notes and transcripts from ACI-bench \citep{aci-bench}, each structured into four standard sections and used for benchmarking and evaluation.
Further details can be found in \secref{asec:data}.

\begin{table}
    \centering
    \small
    \begin{tabular}{ p{0.75\linewidth}   >{\raggedleft\arraybackslash}p{0.15\linewidth} }
        \toprule
        \textbf{Feedback} & \textbf{Rating}  \\
        \cmidrule(lr){1-2}
        ``Does not capture all the key clinical details discussed during the visit'' & 2 \\
        ``Prefer past tense'' & 4  \\
        ``Could benefit from clearer summary and clinical reasoning'' & 3  \\
        ``Covers everything I would have wanted to see'' & 5  \\
        \bottomrule
    \end{tabular}
    \caption{\label{tab:feedback} Example synthetic user feedback and star ratings. 
    } 
\end{table}

\section{Pipeline \& Implementation}
\label{sec:pipeline}

We start with some definitions:

\begin{itemize}[leftmargin=*,itemsep=0em]
    \item {\bf Checklist question}: a question that can be answered with ``Yes'' or ``No''; specifically, the question is designed to be answered with ``Yes'' if the note meets a desired criteria of good clinical note.
    \item {\bf Checklist}: a series of checklist questions that can be used to evaluate a clinical note.
    \item {\bf Passing (a question)}: if a note scores ``Yes'' 
    for the checklist question.
    \item {\bf Checklist score}: for a note and a checklist, the proportion of the checklist questions that pass.
\end{itemize}

To streamline the evaluation process and make checklists more easily comparable, we make several simplifying assumptions:

\begin{itemize}[leftmargin=*,itemsep=0em]
    \item We focus on the Assessment and Plan ({\it AP}) section of the notes, since it is a key component of clinical notes and all encounters should have this section. 
    However, our proposed methods can be extended to other note sections.
    \item Checklist questions must be answerable w/ ``Yes'' or ``No'' for all encounters, i.e., no ``N/A'' answers allowed.
    \item Questions should be answerable given only transcript and the specified note section. %
    \item Checklists should be concise, to be easily processable by humans.

\end{itemize}

\figref{fig:pipeline} shows the proposed pipeline\footnote{This pipeline is illustrative, uses de-identified data, and remains subject to further validation.}.
We will go over generation (\secref{ssec:generation}) and refinement (\secref{ssec:refinement}) of the checklist questions below. See  \secref{sec:eval} for evaluation and results of the final checklist questions.

\subsection{Checklist Generation Methods}
\label{ssec:generation}

We generate candidate checklist questions for each note section separately. Appendix \secref{asec:prompts} contains the prompts used for generating checklists.

\paragraph{Baseline:}

We prompt a LLM with the instruction to generate candidate checklist questions, without any feedback or specified guidelines (zero-shot). This baseline is designed to isolate the contribution of clinician feedback. We chose not to adapt prior rubrics such as \citet{stetson2012assessing} or \citet{10.1136/amiajnl-2013-002321} since they are Likert-style and not directly comparable to binary checklists without careful redesign, and having a clinical expert draft checklist questions from scratch has proven non-trivial and time-consuming, due to the subjectivity and exhaustiveness of the task.

\paragraph{With Feedback:}
We prompt a LLM with the feedback corpus and the instruction to generate candidate checklist questions. Because there is more feedback than can fit in the model's context window, we split the feedback into batches. 
Questions are confirmed to be written such that a ``Yes'' answer corresponds to a good clinical note; if not, the LLM rewrites them in the correct direction.

\subsection{Checklist Refinement}
\label{ssec:refinement}
Once we have candidate checklist questions, we conduct several steps to filter out undesirable questions, based on our criteria in \secref{sec:pipeline}.
The refinement steps, which are detailed below, include:
\begin{enumerate}[leftmargin=*, itemsep=0em]
    \item De-duplicating redundant questions.
    \item Tagging for generally applicable and section specific questions.
    \item Dropping questions that are not LLM enforceable.
    \item Selecting a final subset optimized for feedback coverage and diversity.
\end{enumerate}

\paragraph{Redundant Questions}\label{sssec:redundant}
We obtain embeddings of each question, using \verb|text-embedding-3-large|, and then calculate the cosine similarity between each pair of questions. We build a graph where each question is a node, and an edge exists between two questions if their cosine similarity is $\ge 0.85$ (this threshold is chosen from manual inspection of clusters). 
Each connected component of the graph is a cluster of similar questions, while isolated nodes are unique questions. For each cluster, we prompt \verb|gpt-4o| to generate a single question consolidating the similar questions.

\paragraph{Tagging for applicability and specificity}\label{sssec:tagging}
We keep only questions that are applicable to general encounters, ensuring that they are answerable with a simple ``Yes'' or ``No'' for all encounters, without allowing for ``N/A'' responses. Additionally, we flag questions that reference other note sections, retaining only section-specific questions. This ensures that questions are answerable using only the transcript and the specified note section.
Appendix \secref{assec:flagged_questions} contains examples of questions that would be dropped in this step.
To implement this, we prompt the \verb|o3-mini| model to tag each question using a zero-shot, chain-of-thought (CoT) approach.

\paragraph{LLM Enforceability}
We want to ensure that the questions are answerable by LLM-evaluators. Thus, we define
{\bf enforceability unit tests}: mini-benchmarks of 10 reference notes per checklist question that pass the question criteria, where each reference note is then rewritten to fail the criteria. A unit test is ``passed'' if the rewritten note receives a score of ``No'' for the question. By doing this, we can select the checklist questions that are actually enforceable by LLMs.

We measure the enforceability rate for each checklist question, which is the proportion of rewritten notes that correctly fail the criteria, which is then averaged across all questions into a single \textbf{enforceability score} for the checklist. \secref{assec:enforce} further details the implementation of these unit tests. We discard questions with an enforceability score below a threshold of $0.7$.

\paragraph{Optimal subset of questions for feedback coverage and diversity}
\label{sssec:optimial_select}
The final refinement step selects an optimal subset of checklist questions that maximizes coverage of user feedback while minimizing redundancy and checklist length. We define coverage as the proportion of feedback items addressed by the selected questions, measured using an LLM-based coverage matrix, and diversity as the average dissimilarity (1 minus Jaccard similarity) between the sets of feedback covered by each question. 
Then for coverage $C$, diversity $D$, weight $\alpha$, and length penalty $\lambda$, we can define the following objective score:
\begin{equation*}
    Score(k) = \alpha \cdot C + (1-\alpha) \cdot D - \lambda \cdot k
\end{equation*}
We use beam search to select the subset of $k$ questions that optimizes this score. 
Additional implementation details are provided in \secref{assec:objective}.

\begin{table}
    \centering
    \begin{tabular}{ p{0.43\linewidth}  p{0.2\linewidth}  p{0.2\linewidth}    }
        \toprule
        \textbf{Metric} & \textbf{Baseline ($n=10$)} & \textbf{Feedback ($n=20$)} \\
        \cmidrule(lr){1-3}
        Coverage $\uparrow$ & 0.978 & {\bf 0.987} \\
        Diversity $\uparrow$ & 0.897 & {\bf 0.923} \\
        Intra-cluster Dist $\downarrow$ & 0.670 & {\bf 0.667} \\
        Inter-cluster Dist $\uparrow$ & 0.061 & {\bf 0.080} \\
        LLM enforceability $\uparrow$ & 0.82 & {\bf 0.89} \\
        \bottomrule
    \end{tabular}
    \caption{\label{tab:checklist_eval} Evaluation of the final checklists, with respect to feedback coverage, diversity, and LLM enforceability. Diversity is additionally evaluated by intra-cluster and inter-cluster distances. The feedback checklist beats the baseline checklist in each category. \karen{is updated for new checklist}}
\end{table}

\section{Evaluation}
\label{sec:eval}

Following all refinement steps, we will have a final set of checklist questions. We then evaluate the utility of the final checklist, using the following suite of evaluation metrics:
\begin{itemize}[leftmargin=*,itemsep=0em]
    \item Scores for feedback coverage, feedback diversity, and LLM enforceability.
    \item Predictive power of the checklist questions.
    \item Robustness against perturbations.
    \item Correlation with human preference ratings.
    
\end{itemize}

To demonstrate our pipeline, we generate a checklist from feedback data for the {\it AP} section of the notes. 
From the \textasciitilde$7700$ 
feedback items that are tagged with the {\it AP} section, 97 candidate checklist questions are generated. Following our refinement steps, we obtain a
final checklist of 20 questions. Our checklist is compared against a baseline checklist of 10 questions, both of which are written out in \secref{asec:checklists}. \secref{assec:gen_refine} describes our specific generation and refinement results.

\subsection{Feedback Coverage \& Diversity and LLM Enforceability}
\label{ssec:coverage_and_unit_tests}

Following the definitions given in \secref{ssec:refinement}, we can measure the coverage, diversity, and LLM enforceability of our checklist questions compared to a baseline checklist.
We additionally evaluate diversity by embedding the feedback covered by each question (using \verb|text-embedding-3-large|), and then minimizing intra-cluster distances among questions and maximizing inter-cluster distances between question clusters. Intra-cluster distance is defined as the average distance between all feedback items covered by a question, and inter-cluster distance is defined as the average distance between the centroids of the clusters of feedback items covered by each question.

\para{Our checklist has better feedback coverage and diversity and is more LLM enforceable} than the baseline checklist, as shown in \tbref{tab:checklist_eval}. The feedback checklist has a coverage of 0.987 and a diversity of 0.923, compared to the baseline's coverage of 0.978 and diversity of 0.897. 
The feedback checklist also has a lower intra-cluster distance and a higher inter-cluster distance than the baseline checklist, indicating that it covers a wider range of feedback items while avoiding redundancy.
Our feedback checklist also obtains a higher enforceability score, i.e., it is more answerable by LLMs  (see \secref{assec:eval_metrics} for the breakdown across questions).

\subsection{Predictive Power} 
\label{ssec:shap}
We measure the predictive power of the checklist questions by using them as features to predict the star ratings of the notes. We can do this by training a classifier on the feedback data, where the input is the feedback and the output is the star rating of the note. The checklist questions are used as features for this task, with the mapping of the input to features as the coverage matrix from \secref{sssec:optimial_select}. \secref{assec:eval_metrics} describes the setup of the task in more detail.

\para{Our checklist has better predictive power} than the baseline checklist as well. Our feedback checklist yields a higher accuracy of 0.70 for predicting the star ratings of the notes, compared to 0.62 for the baseline checklist. We also find that our feedback checklist has a higher macro F1 score of 0.63, compared to 0.43 for the baseline checklist. \karen{is updated}
 This indicates that the feedback checklist is better aligned with the star ratings of the notes.

\begin{figure}
    \centering
    \includegraphics[width=0.9\linewidth]{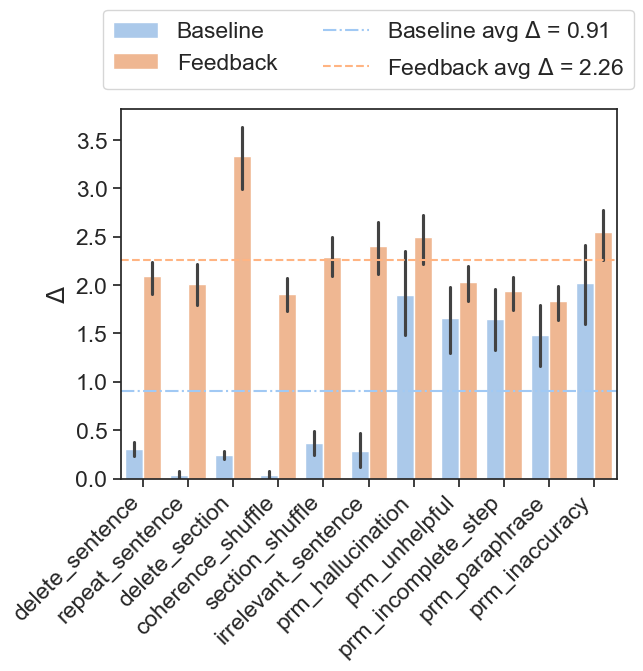}
    \caption{The feedback checklist has a higher perturbation $\Delta$ than the baseline checklist. It is more robust against perturbations, particularly for missing information, organization, and redundancy/hallucination.}
    \label{fig:perturbation_results}
\end{figure}

\subsection{Robustness against Perturbations}
\label{ssec:perturb}

As described in \secref{ssec:aci-bench}, we leverage ACI-bench notes as a set of \textbf{reference notes} that are of good quality. We then define a set of \textbf{perturbations}, i.e., transformations, that can be applied to the reference notes, to simulate poor quality notes. The specific perturbations we apply to the reference notes are described in \secref{assec:eval_metrics} and summarized in \tbref{tab:perturbations}. After scoring the reference notes and the perturbed notes with the checklist, we can measure the robustness of the checklist against these perturbations.

For a checklist question, we define its {\bf Perturbation Delta ($\Delta$)} as the score of the perturbed note subtracted from the score of the reference note, 
averaged over all notes. Since a perturbed note should score lower than the reference note,  $\Delta$ should be larger for more robust checklist questions. We can then measure the robustness of the full checklist by measuring the average $\Delta$ across all questions.

\para{Our checklist is significantly more robust against perturbations} than the baseline checklist. As shown in \figref{fig:perturbation_results}, the feedback checklist has an average perturbation $\Delta$ of 2.26, compared to 0.91 for the baseline checklist. 
The $\Delta$'s are significantly greater than 0 for all perturbations ($p < 0.001$, one-sample $t$-test). In contrast, the baseline shows no significant change for the \verb|repeat_sentence| and \verb|coherence_shuffle| perturbations. We also confirm that most $\Delta$'s are significantly larger for our feedback checklist than the baseline, with $p<0.05$ for a two-sample $t$-test and a large effect size (Cohen's $d=2.94$).

Our checklist is particularly robust to perturbations like \verb|delete_sentence| and \verb|delete_section|, which are designed to simulate missing salient information.
 Furthermore, our checklist is more sensitive to perturbations like \verb|section_shuffle| and \verb|coherence_shuffle|, which are designed to simulate poor writing flow and organization. 
\verb|repeat_sentence| and \verb|irrelevant_sentence| are also detected by our feedback checklist, but not by the baseline checklist, suggesting that our feedback checklist is better at detecting redundancy and hallucination errors in the notes.

\subsection{Correlation with Human Preference Ratings}
\label{ssec:correlation}

We use $109$ 
unique pairs of preference ratings for the {\it AP} section for our evaluation. 
For each note pair, we compute the checklist score for each note. We then aggregate the scores for the ranked notes and check for statistical significance. Ideally, the preferred note should have a higher checklist score than the non-preferred note.

\begin{figure}
    \centering
    \includegraphics[width=0.95\linewidth]{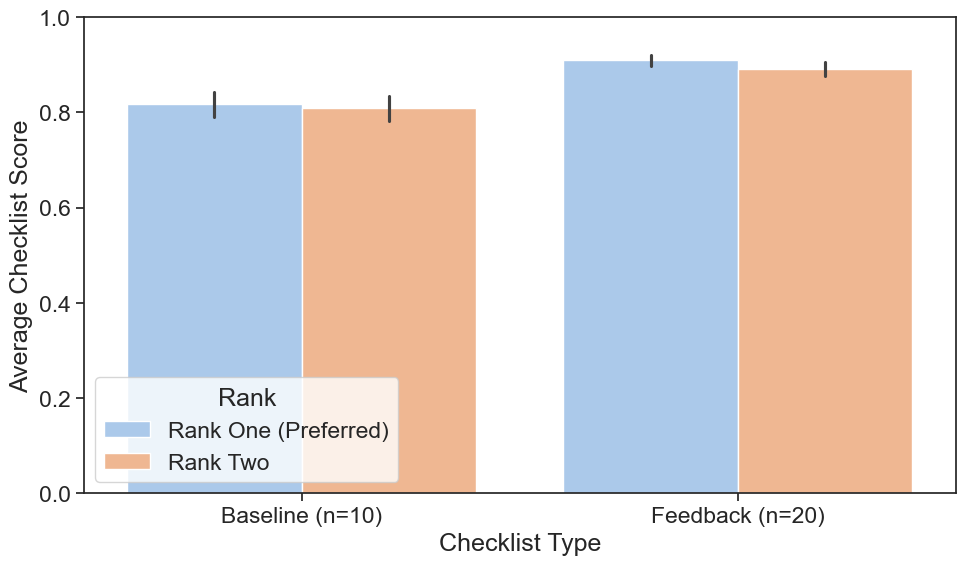}
    \caption{Correlation with human preference ratings is significant for our checklist ($p \le 0.05$ from a paired $t$-test, Cohen's $d=0.28$), but not for the baseline. }
    \label{fig:eyes}
\end{figure}

\para{Our checklist is significantly correlated with human preferences} compared to the baseline checklist (see \figref{fig:eyes}). That is, our checklist produces a significant difference ($p \le 0.05$ from a paired $t$-test) in scores between the preferred and non-preferred notes, with prefered notes having the higher score. The baseline checklist does not produce a significant difference in scores between the preferred vs. non-preferred notes.

\section{Conclusion}
\label{sec:conclusion}

In this work, we propose a method to automatically generate a checklist for evaluating clinical notes, based on real user feedback for notes produced by a commercial Ambient AI company.
We show that our generated checklists have more desirable properties than a baseline checklist, 
and we define a series of evaluation metrics to assess the quality of the checklists. 
In particular, we show that our checklists are comprehensive, diverse, predictive of human ratings, and aligned with expert preferences.

Our approach addresses a critical need in healthcare AI for scalable, clinically-grounded evaluation. By systematically distilling feedback from over 21,000 real clinical encounters, we demonstrate that user concerns can be transformed into actionable evaluation criteria. The resulting checklist serves as an interpretable, enforceable tool that can flag potentially problematic notes while maintaining transparency about evaluation criteria.

This work demonstrates the practical value of leveraging large-scale user feedback to create evaluation tools for deployed AI systems. Our pipeline provides a systematic approach to convert real user feedback into structured evaluation criteria, offering a scalable alternative to purely manual review while preserving clinical input and oversight.

Future work includes scaling the pipeline to additional note sections and domains, implementing dynamic and more robust feedback filtering, and incorporating advanced evaluation methods such as feature importance analysis, human studies, and improved LLM-evaluator reasoning to further refine and validate the checklists. 
Extended discussion of future work can be found in \secref{asec:future}.

\section*{Limitations} 
\label{sec:limitations}

Our pipeline makes several simplifying assumptions, such as excluding "N/A" answers and cross-section comparisons, which may limit the scope and nuance of the resulting checklists. 
Checklist coverage is inherently limited by the content of user feedback; e.g., if there are no complaints about pronoun usage, there may not be a checklist question about using correct pronouns.  
Checklist questions that favor comprehensive notes may penalize concise ones that are still clinically sufficient, such as in time-sensitive settings like emergency care. Our methodology could be adapted for these specialties by filtering for specialty-specific feedback, as well as for other contexts (e.g., multi-provider notes).

Our current setup treats all checklist questions with equal weight, and though this assessment is inherently subjective, we acknowledge that clinical significance varies across checklist items.  Weighted scoring could be implemented automatically (e.g., \citet{wei2025rocketeval}) or added during clinical review of the rubric as a post-hoc step, without redesigning our existing methodology.

We rely heavily on LLMs in our pipeline and evaluation, which introduces its own set of challenges. When using LLMs-as-a-Judge, there is a lack of ground truth labels to determine whether clinical notes satisfy checklist questions, such as for ACI-bench notes. 
Exhaustive evaluation is difficult, since each checklist question can be a substantial problem on its own. As such, our current metrics are not exhaustive.
Finally, we rely on proxies for positive signals, such as internal pairwise preference data, though these proxies also have their own limitations (e.g., poor agreement between raters, though we mitigate this by selecting pairs with at least two raters in agreement).

\section*{Ethical Considerations}
\label{sec:ethics}
\karen{not counted in page limit}

Our work involves the use of clinical note data and user feedback for the purpose of building automated evaluation pipelines. All proprietary data used in this work, such as user feedback, star ratings, and clinical note content, is handled in compliance with applicable data privacy policies and HIPAA regulations. All PHI is de-identified in accordance with HIPAA safe harbor standard before model training and evaluation; no direct identifiers are transmitted outside the covered environment.

Our checklist is designed as a scalable triage tool, not a replacement for clinical judgment.  Our generated checklists are designed to be interpretable and transparent, but further validation with practicing clinicians is necessary before clinical deployment.
 Clinicians can review the finite set of criteria, after which the checklist can be applied at scale to flag notes for expert attention. 
We acknowledge that LLM outputs may introduce subtle errors or biases, and we flag this as an area for continued validation.
In production, the checklist will be part of a much larger evaluation suite that involves human and clinician oversight at different stages, not to be used as a single criterion for decision making. 

\section*{Acknowledgments}
We thank the anonymous reviewers for their comments, which helped improve this work. 
We also thank Zack Lipton, Elisa Ferracane-Kelley, Kaveh Moghbeli, Mike Myerburg, and the broader clinical and machine learning teams at Abridge, for their support and feedback throughout this project.

\bibliography{custom}  %

\begin{thebibliography}{39}
\providecommand{\natexlab}[1]{#1}

\bibitem[{Amenyo et~al.(2025)Amenyo, Grossman, Brown, and
  Wylie-Toal}]{amenyo2025assessmentaigeneratedpediatricrehabilitation}
Solomon Amenyo, Maura~R. Grossman, Daniel~G. Brown, and Brendan Wylie-Toal.
  2025.
\newblock \href {https://arxiv.org/abs/2503.15526} {Assessment of ai-generated
  pediatric rehabilitation soap-note quality}.
\newblock \emph{Preprint}, arXiv:2503.15526.

\bibitem[{Brykczynski(1999)}]{10.1145/308769.308798}
Bill Brykczynski. 1999.
\newblock \href {https://doi.org/10.1145/308769.308798} {A survey of software
  inspection checklists}.
\newblock \emph{SIGSOFT Softw. Eng. Notes}, 24(1):82.

\bibitem[{Burke et~al.(2014)Burke, Hoang, Becher, Fontelo, Liu, Stephens,
  Pangaro, Sessums, O'Malley, Baxi, Bunt, Capaldi, Chen, Cooper, Djuric, Hodge,
  Kane, Magee, Makary, Mallory, Miller, Saperstein, Servey, and
  Gimbel}]{10.1136/amiajnl-2013-002321}
Harry~B Burke, Albert Hoang, Dorothy Becher, Paul Fontelo, Fang Liu, Mark
  Stephens, Louis~N Pangaro, Laura~L Sessums, Patrick O'Malley, Nancy~S Baxi,
  Christopher~W Bunt, Vincent~F Capaldi, Julie~M Chen, Barbara~A Cooper,
  David~A Djuric, Joshua~A Hodge, Shawn Kane, Charles Magee, Zizette~R Makary,
  Renee~M Mallory, Thomas Miller, Adam Saperstein, Jessica Servey, and Ronald~W
  Gimbel. 2014.
\newblock \href {https://doi.org/10.1136/amiajnl-2013-002321} {Qnote: an
  instrument for measuring the quality of ehr clinical notes}.
\newblock \emph{Journal of the American Medical Informatics Association},
  21(5):910--916.

\bibitem[{Chaparro et~al.(2019)Chaparro, Keebler, Lazzara, and
  Diamond}]{Chaparro2019ChecklistsAR}
Alex Chaparro, Joseph~Roland Keebler, Elizabeth~H. Lazzara, and Anastasia
  Diamond. 2019.
\newblock \href
  {https://journals.sagepub.com/doi/epdf/10.1177/1064804618819181} {Checklists:
  A review of their origins, benefits, and current uses as a cognitive aid in
  medicine}.
\newblock \emph{Ergonomics in Design: The Quarterly of Human Factors
  Applications}, 27:21 -- 26.

\bibitem[{Chu et~al.(2025)Chu, Kim, and Yi}]{chu2025thinkworkbettercombining}
SeongYeub Chu, JongWoo Kim, and MunYong Yi. 2025.
\newblock \href {https://arxiv.org/abs/2409.07355} {Think together and work
  better: Combining humans' and llms' think-aloud outcomes for effective text
  evaluation}.
\newblock \emph{Preprint}, arXiv:2409.07355.

\bibitem[{Cook et~al.(2024)Cook, Rocktäschel, Foerster, Aumiller, and
  Wang}]{cook2024tickingboxesgeneratedchecklists}
Jonathan Cook, Tim Rocktäschel, Jakob Foerster, Dennis Aumiller, and Alex
  Wang. 2024.
\newblock \href {https://arxiv.org/abs/2410.03608} {Ticking all the boxes:
  Generated checklists improve llm evaluation and generation}.
\newblock \emph{Preprint}, arXiv:2410.03608.

\bibitem[{Croxford et~al.(2025{\natexlab{a}})Croxford, Gao, Pellegrino, Wong,
  Wills, First, Liao, Goswami, Patterson, and Afshar}]{Croxford2025CurrentAF}
Emma Croxford, Yanjun Gao, Nicholas Pellegrino, Karen~K. Wong, Graham Wills,
  Elliot First, Frank~J Liao, Cherodeep Goswami, Brian~W. Patterson, and Majid
  Afshar. 2025{\natexlab{a}}.
\newblock \href {https://api.semanticscholar.org/CorpusID:276140921} {Current
  and future state of evaluation of large language models for medical
  summarization tasks}.
\newblock \emph{Npj health systems}, 2.

\bibitem[{Croxford et~al.(2025{\natexlab{b}})Croxford, Gao, Pellegrino, Wong,
  Wills, First, Schnier, Burton, Ebby, Gorskic, Kalscheur, Khalil, Pisani,
  Rubeor, Stetson, Liao, Goswami, Patterson, and
  Afshar}]{croxford2025developmentvalidationproviderdocumentation}
Emma Croxford, Yanjun Gao, Nicholas Pellegrino, Karen~K. Wong, Graham Wills,
  Elliot First, Miranda Schnier, Kyle Burton, Cris~G. Ebby, Jillian Gorskic,
  Matthew Kalscheur, Samy Khalil, Marie Pisani, Tyler Rubeor, Peter Stetson,
  Frank Liao, Cherodeep Goswami, Brian Patterson, and Majid Afshar.
  2025{\natexlab{b}}.
\newblock \href {https://arxiv.org/abs/2501.08977} {Development and validation
  of the provider documentation summarization quality instrument for large
  language models}.
\newblock \emph{Preprint}, arXiv:2501.08977.

\bibitem[{Degani and Wiener(1993)}]{doi:10.1177/001872089303500209}
Asaf Degani and Earl~L. Wiener. 1993.
\newblock \href {https://doi.org/10.1177/001872089303500209} {Cockpit
  checklists: Concepts, design, and use}.
\newblock \emph{Human Factors}, 35(2):345--359.

\bibitem[{Eng et~al.(2024)Eng, Johnston, Cerda, Kadakia, Mosier-Mills, and
  Vanka}]{Eng2024APD}
Kathleen Eng, Katherine~T Johnston, Ivo~H. Cerda, Kushal~T. Kadakia, Alison
  Mosier-Mills, and Anita Vanka. 2024.
\newblock A patient-centered documentation skills curriculum for preclerkship
  medical students in an open notes era.
\newblock \emph{MedEdPORTAL : the Journal of Teaching and Learning Resources},
  20.

\bibitem[{Goldberg et~al.(2024)Goldberg, Ullah, Khuong, Rachmat, Xu, Guyon, and
  Shah}]{goldberg2024usefulness}
Alexander Goldberg, Ihsan Ullah, Thanh Gia~Hieu Khuong, Benedictus~Kent
  Rachmat, Zhen Xu, Isabelle Guyon, and Nihar~B Shah. 2024.
\newblock Usefulness of llms as an author checklist assistant for scientific
  papers: Neurips'24 experiment.
\newblock \emph{arXiv preprint arXiv:2411.03417}.

\bibitem[{Guo et~al.(2024)Guo, August, Leroy, Cohen, and
  Wang}]{guo2024applsevaluatingevaluationmetrics}
Yue Guo, Tal August, Gondy Leroy, Trevor Cohen, and Lucy~Lu Wang. 2024.
\newblock \href {https://arxiv.org/abs/2305.14341} {Appls: Evaluating
  evaluation metrics for plain language summarization}.
\newblock \emph{Preprint}, arXiv:2305.14341.

\bibitem[{Hanson et~al.(2012)Hanson, Stephens, Pangaro, and
  Gimbel}]{Hanson2012QualityOO}
Janice~L. Hanson, Mark~B. Stephens, Louis~N. Pangaro, and Ronald~W. Gimbel.
  2012.
\newblock Quality of outpatient clinical notes: a stakeholder definition
  derived through qualitative research.
\newblock \emph{BMC Health Services Research}, 12:407 -- 407.

\bibitem[{Haynes et~al.(2009)Haynes, Weiser, Berry, Lipsitz, Breizat,
  Dellinger, Herbosa, Joseph, Kibatala, Lapitan, Merry, Moorthy, Reznick,
  Taylor, and Gawande}]{doi:10.1056/NEJMsa0810119}
Alex~B. Haynes, Thomas~G. Weiser, William~R. Berry, Stuart~R. Lipsitz,
  Abdel-Hadi~S. Breizat, E.~Patchen Dellinger, Teodoro Herbosa, Sudhir Joseph,
  Pascience~L. Kibatala, Marie Carmela~M. Lapitan, Alan~F. Merry, Krishna
  Moorthy, Richard~K. Reznick, Bryce Taylor, and Atul~A. Gawande. 2009.
\newblock \href {https://doi.org/10.1056/NEJMsa0810119} {A surgical safety
  checklist to reduce morbidity and mortality in a global population}.
\newblock \emph{New England Journal of Medicine}, 360(5):491--499.

\bibitem[{Kanithi et~al.(2024)Kanithi, Christophe, Pimentel, Raha, Saadi,
  Javed, Maslenkova, Hayat, Rajan, and
  Khan}]{kanithi2024mediccomprehensiveframeworkevaluating}
Praveen~K Kanithi, Clément Christophe, Marco~AF Pimentel, Tathagata Raha, Nada
  Saadi, Hamza Javed, Svetlana Maslenkova, Nasir Hayat, Ronnie Rajan, and
  Shadab Khan. 2024.
\newblock \href {https://arxiv.org/abs/2409.07314} {Medic: Towards a
  comprehensive framework for evaluating llms in clinical applications}.
\newblock \emph{Preprint}, arXiv:2409.07314.

\bibitem[{Lee et~al.(2025)Lee, Kim, Kim, Cho, Kang, Kang, and
  Kim}]{lee2025checkevalreliablellmasajudgeframework}
Yukyung Lee, Joonghoon Kim, Jaehee Kim, Hyowon Cho, Jaewook Kang, Pilsung Kang,
  and Najoung Kim. 2025.
\newblock \href {https://arxiv.org/abs/2403.18771} {Checkeval: A reliable
  llm-as-a-judge framework for evaluating text generation using checklists}.
\newblock \emph{Preprint}, arXiv:2403.18771.

\bibitem[{Li et~al.(2025)Li, Li, and
  Tan}]{li2025hypoevalhypothesisguidedevaluationnatural}
Mingxuan Li, Hanchen Li, and Chenhao Tan. 2025.
\newblock \href {https://arxiv.org/abs/2504.07174} {Hypoeval: Hypothesis-guided
  evaluation for natural language generation}.
\newblock \emph{Preprint}, arXiv:2504.07174.

\bibitem[{Lin et~al.(2024)Lin, Deng, Chandu, Brahman, Ravichander, Pyatkin,
  Dziri, Bras, and Choi}]{lin2024wildbenchbenchmarkingllmschallenging}
Bill~Yuchen Lin, Yuntian Deng, Khyathi Chandu, Faeze Brahman, Abhilasha
  Ravichander, Valentina Pyatkin, Nouha Dziri, Ronan~Le Bras, and Yejin Choi.
  2024.
\newblock \href {https://arxiv.org/abs/2406.04770} {Wildbench: Benchmarking
  llms with challenging tasks from real users in the wild}.
\newblock \emph{Preprint}, arXiv:2406.04770.

\bibitem[{Lundberg and Lee(2017)}]{NIPS2017_7062}
Scott~M Lundberg and Su-In Lee. 2017.
\newblock \href
  {http://papers.nips.cc/paper/7062-a-unified-approach-to-interpreting-model-predictions.pdf}
  {A unified approach to interpreting model predictions}.
\newblock In I.~Guyon, U.~V. Luxburg, S.~Bengio, H.~Wallach, R.~Fergus,
  S.~Vishwanathan, and R.~Garnett, editors, \emph{Advances in Neural
  Information Processing Systems 30}, pages 4765--4774. Curran Associates, Inc.

\bibitem[{Nan et~al.(2017)Nan, van Gorp, Lu, Kaymak, Korsten, Vdovjak, and
  Duan}]{Nan2017AMF}
Shan Nan, Pieter van Gorp, Xudong Lu, Uzay Kaymak, Hendrikus H.~M. Korsten,
  Richard Vdovjak, and Huilong Duan. 2017.
\newblock \href
  {https://bmcmedinformdecismak.biomedcentral.com/articles/10.1186/s12911-017-0551-0}
  {A meta-model for computer executable dynamic clinical safety checklists}.
\newblock \emph{BMC Medical Informatics and Decision Making}, 17.

\bibitem[{Oleson(2024)}]{oleson2024deepscorecomprehensiveapproachmeasuring}
Jon Oleson. 2024.
\newblock \href {https://arxiv.org/abs/2409.16307} {Deepscore: A comprehensive
  approach to measuring quality in ai-generated clinical documentation}.
\newblock \emph{Preprint}, arXiv:2409.16307.

\bibitem[{Pereira et~al.(2024)Pereira, Assumpcao, and
  Lotufo}]{pereira2024checkevalchecklistbasedapproachevaluating}
Jayr Pereira, Andre Assumpcao, and Roberto Lotufo. 2024.
\newblock \href {https://arxiv.org/abs/2407.14467} {Check-eval: A
  checklist-based approach for evaluating text quality}.
\newblock \emph{Preprint}, arXiv:2407.14467.

\bibitem[{Que et~al.(2024)Que, Duan, He, Mou, Zhou, Liu, Rong, Wang, Yang,
  Zhang, Peng, Zhang, Zhang, and Chen}]{que2024hellobenchevaluatinglongtext}
Haoran Que, Feiyu Duan, Liqun He, Yutao Mou, Wangchunshu Zhou, Jiaheng Liu,
  Wenge Rong, Zekun~Moore Wang, Jian Yang, Ge~Zhang, Junran Peng, Zhaoxiang
  Zhang, Songyang Zhang, and Kai Chen. 2024.
\newblock \href {https://arxiv.org/abs/2409.16191} {Hellobench: Evaluating long
  text generation capabilities of large language models}.
\newblock \emph{Preprint}, arXiv:2409.16191.

\bibitem[{Reijers et~al.(2017)Reijers, Leopold, and
  Recker}]{Reijers2017TowardsAS}
Hajo~Alexander Reijers, Henrik Leopold, and Jan Recker. 2017.
\newblock Towards a science of checklists.
\newblock In \emph{Hawaii International Conference on System Sciences}.

\bibitem[{Savkov et~al.(2022)Savkov, Moramarco, Papadopoulos~Korfiatis, Perera,
  Belz, and Reiter}]{savkov-etal-2022-consultation}
Aleksandar Savkov, Francesco Moramarco, Alex Papadopoulos~Korfiatis, Mark
  Perera, Anya Belz, and Ehud Reiter. 2022.
\newblock \href {https://doi.org/10.18653/v1/2022.emnlp-industry.10}
  {Consultation checklists: Standardising the human evaluation of medical note
  generation}.
\newblock In \emph{Proceedings of the 2022 Conference on Empirical Methods in
  Natural Language Processing: Industry Track}, pages 111--120, Abu Dhabi, UAE.
  Association for Computational Linguistics.

\bibitem[{Stetson et~al.(2012)Stetson, Bakken, Wrenn, and
  Siegler}]{stetson2012assessing}
Peter~D Stetson, Suzanne Bakken, Jesse~O Wrenn, and Eugenia~L Siegler. 2012.
\newblock Assessing electronic note quality using the physician documentation
  quality instrument (pdqi-9).
\newblock \emph{Applied clinical informatics}, 3(02):164--174.

\bibitem[{Stetson et~al.(2008)Stetson, Morrison, Bakken, and
  Johnson}]{STETSON2008534}
Peter~D. Stetson, Frances~P. Morrison, Suzanne Bakken, and Stephen~B. Johnson.
  2008.
\newblock \href {https://doi.org/10.1197/jamia.M2404} {Preliminary development
  of the physician documentation quality instrument}.
\newblock \emph{Journal of the American Medical Informatics Association},
  15(4):534--541.

\bibitem[{Wang et~al.(2025{\natexlab{a}})Wang, Gao, Liu, Xu, Hussein, Labban,
  Iheasirim, Korsapati, Outcalt, and
  Sun}]{wang2025adaptingopensourcelargelanguage}
Hanyin Wang, Chufan Gao, Bolun Liu, Qiping Xu, Guleid Hussein, Mohamad~El
  Labban, Kingsley Iheasirim, Hariprasad Korsapati, Chuck Outcalt, and Jimeng
  Sun. 2025{\natexlab{a}}.
\newblock \href {https://arxiv.org/abs/2405.00715} {Towards adapting
  open-source large language models for expert-level clinical note generation}.
\newblock \emph{Preprint}, arXiv:2405.00715.

\bibitem[{Wang et~al.(2025{\natexlab{b}})Wang, Gao, Xu, Liu, Hussein,
  Korsapati, Labban, Iheasirim, Hassan, Anil, Bartlett, and
  Sun}]{wang2025processsupervisedrewardmodelsverifying}
Hanyin Wang, Chufan Gao, Qiping Xu, Bolun Liu, Guleid Hussein, Hariprasad
  Korsapati, Mohamad~El Labban, Kingsley Iheasirim, Mohamed Hassan, Gokhan
  Anil, Brian Bartlett, and Jimeng Sun. 2025{\natexlab{b}}.
\newblock \href {https://arxiv.org/abs/2412.12583} {Process-supervised reward
  models for verifying clinical note generation: A scalable approach guided by
  domain expertise}.
\newblock \emph{Preprint}, arXiv:2412.12583.

\bibitem[{Wang et~al.(2024)Wang, Wang, Danek, Li, Mack, Poon, Wang, Rajpurkar,
  and Sun}]{wang2024perspectiveadaptinggeneralistai}
Zifeng Wang, Hanyin Wang, Benjamin Danek, Ying Li, Christina Mack, Hoifung
  Poon, Yajuan Wang, Pranav Rajpurkar, and Jimeng Sun. 2024.
\newblock \href {https://arxiv.org/abs/2411.00024} {A perspective for adapting
  generalist ai to specialized medical ai applications and their challenges}.
\newblock \emph{Preprint}, arXiv:2411.00024.

\bibitem[{Wang et~al.(2023)Wang, Shang, and Zhong}]{wang-etal-2023-goal}
Zihan Wang, Jingbo Shang, and Ruiqi Zhong. 2023.
\newblock \href {https://doi.org/10.18653/v1/2023.emnlp-main.657} {Goal-driven
  explainable clustering via language descriptions}.
\newblock In \emph{Proceedings of the 2023 Conference on Empirical Methods in
  Natural Language Processing}, pages 10626--10649, Singapore. Association for
  Computational Linguistics.

\bibitem[{Wei et~al.(2025)Wei, Wen, Qiao, Sun, and Ma}]{wei2025rocketeval}
Tianjun Wei, Wei Wen, Ruizhi Qiao, Xing Sun, and Jianghong Ma. 2025.
\newblock \href {https://openreview.net/forum?id=zJjzNj6QUe} {Rocketeval:
  Efficient automated {LLM} evaluation via grading checklist}.
\newblock In \emph{The Thirteenth International Conference on Learning
  Representations}.

\bibitem[{Xie et~al.(2024)Xie, Zhang, Cheng, Liu, Gero, Wong, Naumann, Poon,
  and Rose}]{xie-etal-2024-doclens}
Yiqing Xie, Sheng Zhang, Hao Cheng, Pengfei Liu, Zelalem Gero, Cliff Wong,
  Tristan Naumann, Hoifung Poon, and Carolyn Rose. 2024.
\newblock \href {https://doi.org/10.18653/v1/2024.acl-long.39} {{D}oc{L}ens:
  Multi-aspect fine-grained evaluation for medical text generation}.
\newblock In \emph{Proceedings of the 62nd Annual Meeting of the Association
  for Computational Linguistics (Volume 1: Long Papers)}, pages 649--679,
  Bangkok, Thailand. Association for Computational Linguistics.

\bibitem[{Ye et~al.(2024)Ye, Wang, Huang, Chen, Zhang, Moniz, Gao, Geyer,
  Huang, Chen et~al.}]{ye2024justice}
Jiayi Ye, Yanbo Wang, Yue Huang, Dongping Chen, Qihui Zhang, Nuno Moniz, Tian
  Gao, Werner Geyer, Chao Huang, Pin-Yu Chen, et~al. 2024.
\newblock Justice or prejudice? quantifying biases in llm-as-a-judge.
\newblock \emph{arXiv preprint arXiv:2410.02736}.

\bibitem[{Yim et~al.(2023)Yim, Fu, {Ben Abacha}, Snider, Lin, and
  Yetisgen}]{aci-bench}
Wen{-}wai Yim, Yujuan Fu, Asma {Ben Abacha}, Neal Snider, Thomas Lin, and
  Meliha Yetisgen. 2023.
\newblock Aci-bench: a novel ambient clinical intelligence dataset for
  benchmarking automatic visit note generation.
\newblock \emph{Nature Scientific Data}.

\bibitem[{Yim et~al.(2019)Yim, Mills, Chun, Hashiguchi, Yew, and
  Lu}]{yim-etal-2019-automatic}
Wen-wai Yim, Ashley Mills, Harold Chun, Teresa Hashiguchi, Justin Yew, and
  Bryan Lu. 2019.
\newblock \href {https://doi.org/10.18653/v1/D19-6216} {Automatic rubric-based
  content grading for clinical notes}.
\newblock In \emph{Proceedings of the Tenth International Workshop on Health
  Text Mining and Information Analysis (LOUHI 2019)}, pages 126--135, Hong
  Kong. Association for Computational Linguistics.

\bibitem[{Zhang et~al.(2021)Zhang, Morris, Ustun, and
  Ghassemi}]{10.5555/3540261.3540355}
Haoran Zhang, Quaid Morris, Berk Ustun, and Maryzeh Ghassemi. 2021.
\newblock Learning optimal predictive checklists.
\newblock In \emph{Proceedings of the 35th International Conference on Neural
  Information Processing Systems}, NIPS '21, Red Hook, NY, USA. Curran
  Associates Inc.

\bibitem[{Zhang et~al.(2025)Zhang, Shen, Li, Sha, Hu, Wang, Huang, Liu, Tong,
  Jiang, Chai, Xi, Dou, Gui, Zhang, and
  Huang}]{zhang2025llmevalmedrealworldclinicalbenchmark}
Ming Zhang, Yujiong Shen, Zelin Li, Huayu Sha, Binze Hu, Yuhui Wang, Chenhao
  Huang, Shichun Liu, Jingqi Tong, Changhao Jiang, Mingxu Chai, Zhiheng Xi,
  Shihan Dou, Tao Gui, Qi~Zhang, and Xuanjing Huang. 2025.
\newblock \href {https://arxiv.org/abs/2506.04078} {Llmeval-med: A real-world
  clinical benchmark for medical llms with physician validation}.
\newblock \emph{Preprint}, arXiv:2506.04078.

\bibitem[{Zheng et~al.(2023)Zheng, Chiang, Sheng, Zhuang, Wu, Zhuang, Lin, Li,
  Li, Xing et~al.}]{zheng2023judging}
Lianmin Zheng, Wei-Lin Chiang, Ying Sheng, Siyuan Zhuang, Zhanghao Wu, Yonghao
  Zhuang, Zi~Lin, Zhuohan Li, Dacheng Li, Eric Xing, et~al. 2023.
\newblock Judging llm-as-a-judge with mt-bench and chatbot arena.
\newblock \emph{Advances in Neural Information Processing Systems},
  36:46595--46623.

\end{thebibliography}

\appendix

\section{Additional Related Work}
\label{asec:related_work}

\paragraph{Checklist generation} 
With the utility of checklists being proven across high-stakes domains \citep{Reijers2017TowardsAS, Chaparro2019ChecklistsAR, 10.1145/308769.308798, doi:10.1177/001872089303500209, doi:10.1056/NEJMsa0810119}, there is a growing body of work on generating checklists automatically with LLMs \citep{lee2025checkevalreliablellmasajudgeframework, cook2024tickingboxesgeneratedchecklists, pereira2024checkevalchecklistbasedapproachevaluating, wei2025rocketeval, li2025hypoevalhypothesisguidedevaluationnatural}. These works focus on generating checklists specific to individual instructions or data samples (e.g., a summary or user conversation), rather than generalizable checklists that can be applied across a corpus of data.
\chenhao{not sure what the final sentence meant, the methods in the next paragraph look like working for a corpus?} \karen{like their checklists are specific to the input data, e.g., a summary or user conversation. We want to generate checklists that can be applied across a corpus of clinical notes, not just one note or one summary.}

For other clinical applications, \citet{Nan2017AMF} propose a rule-based method for generating dynamic clinical safety checklists and 
 \citet{10.5555/3540261.3540355} learns optimal checklists for clinical decision support from a set of predefined binarized features.
Checklists have also been manually developed by experts, for individual medical notes \citep{savkov-etal-2022-consultation} and generative medical tasks in Chinese \citep{zhang2025llmevalmedrealworldclinicalbenchmark}.  Our work aims to automatically generate checklists for clinical notes that apply across encounters, checking for items beyond facts.

\paragraph{Checklist evaluation}
\citet{wang-etal-2023-goal} propose a method of generating natural language cluster labels based on a goal; we similarly want to generate checklist questions that cover a corpus of feedback, based on the goal of evaluating note quality. Our feedback coverage methodology is loosely inspired by their Propose-Assign-Select framework.
\citet{guo2024applsevaluatingevaluationmetrics} define a testbed of global perturbations to assess metrics for plain language summarization; we draw from these criteria to form our own perturbation benchmark.

\section{Data}
\label{asec:data}

\subsection{User Feedback}
\label{ssec:feedback}

We retrieve 
\textasciitilde22K
encounters with user feedback of $>2$ words, non-null (1-5) star ratings.
\tbref{tab:feedback} provides some synthetic examples.
The user feedback is free-text, and the star ratings are on a scale of 1-5, with 1 being the worst and 5 being the best; these are obtained from real users of an Ambient AI company that scribes clinical encounters and generates clinical notes. 
We specifically filter for feedback that is relevant to note quality and content, as opposed to other issues like recording or EHR integration.
We use an internal tool to determine the relevant note section (e.g. {\it HPI}, {\it AP}) for each piece of feedback and only keep the relevant feedback for each section.

\subsection{Human Preference Ratings}
\label{ssec:ratings}
We also have access to $506$ pairs of note sections with internal preference ratings, which are rated by clinical experts within the company. These pairs of notes have at least two raters agreeing on the preferred note in a pairwise comparison.
Each sample of this dataset contains the transcript of the encounter and the pair of notes for the section being rated, with the preferred note being marked.

\subsection{Reference Notes: ACI-bench}
\label{ssec:aci-bench}
We are using $207$ reference notes and transcripts from ACI-bench \citep{aci-bench}.
Each note contains four sections, corresponding to the sections our user feedback is tagged with in parentheses:
\begin{itemize}[leftmargin=*, itemsep=0em]
    \item SUBJECTIVE ({\it HPI\_Subjective}) includes items taken during verbal exam and typically written in the chief complaint, history of present illness, and past social history. 
    \item OBJECTIVE EXAM (Physical Exam or {\it PE}) includes content from the physical examination on the day of the visit. 
    \item OBJECTIVE RESULTS ({\it Results}) includes diagnostics taken prior to the visit, including laboratory or imaging results. 
    \item ASSESSMENT AND PLAN ({\it AP}) includes the doctor’s diagnosis and planned tests and treatments. 
\end{itemize}

In cases where certain sections are missing, an EMPTY flag is used as the content.
We utilize these reference notes to calculate pass rates/checklist scores, generate perturbed notes, and create unit tests.

\section{LLM Prompts}
\label{asec:prompts}
\verb|gpt-4o| is used for checklist generation. 

\subsection{Baselines}
\label{assec:baselines}

System message:
\begin{lstlisting}[
  basicstyle=\ttfamily\small, %
  breaklines=true,            %
  frame=single,               %
  columns=flexible,           %
  keepspaces=true,            %
  tabsize=4                   %
]
You are an expert in clinical documentation. You want to evaluate the quality of a clinical note. You will generate a list of simple yes/no questions, such that the "Yes" answer corresponds to a good clinical note. 
You may optionally specify the specific note section that each question pertains to. The sections are as follows:
- "subjective" includes items taken during verbal exam and typically written in the chief complaint, history of present illness, and past social history
- "objective_exam" includes content from the physical examination on the day of the visit 
- "objective_results" includes diagnostics taken prior to the visit, including laboratory or imaging results 
- "assessment_and_plan" includes the doctor's diagnosis and planned tests and treatments
If the question is applicable to the full note, then denote "full" as the pertinent section.
\end{lstlisting}

\subsection{Feedback Checklists}
\label{assec:feedback}
Proposer's system message for a given section:
\begin{lstlisting}[
  basicstyle=\ttfamily\small, %
  breaklines=true,            %
  frame=single,               %
  columns=flexible,           %
  keepspaces=true,            %
  tabsize=4                   %
]
You are an expert in clinical documentation. 
You will be provided with an itemized list of physician feedback on the {section} section of medical notes. 
Your task is to generate a set of yes/no questions that comprehensively reflect this feedback. 

The questions should aim to identify the presence or absence of good documentation practices in the {section} section. 
The questions should be specific to the {section} section and should not reference or depend on the content of other sections of the medical note. 
If a feedback item pertains to multiple note sections, focus only on the parts that apply to {section} content.
Each question should ideally address multiple feedback items rather than targeting only one. 
Keep questions concise, atomic, and objective; avoid complex sentence structure. 
For each question, list the corresponding INDICES of the feedback item that are addressed. 

Avoid overly specific questions; instead, favor general questions that apply across a variety of medical encounters. 
Do not include protected health information (PHI) in any questions.

Use third-person language when referring to the note. 
A "Yes" answer should indicate that the note meets good clinical documentation standards. 

Incorporate both positive and negative feedback: 
- For negative feedback, generate questions that would flag a note if it exhibited similar issues. 
- For positive feedback, generate questions that promote those qualities in all notes. 

Generate as many questions as necessary to cover the feedback comprehensively.
\end{lstlisting}

Assigner's system message:
\begin{lstlisting}[
  basicstyle=\ttfamily\small, %
  breaklines=true,            %
  frame=single,               %
  columns=flexible,           %
  keepspaces=true,            %
  tabsize=4                   %
]
You will be given an itemized checklist of questions and a user feedback. 
Indicate which questions cover the feedback. Respond as a list of question item numbers.

Questions: 
{itemized checklist questions}
\end{lstlisting}

\subsection{Unit Tests}
\label{assec:ut}

System message for re-writing reference notes to fail checklist criteria:
\begin{lstlisting}[
  basicstyle=\ttfamily\small, %
  breaklines=true,            %
  frame=single,               %
  columns=flexible,           %
  keepspaces=true,            %
  tabsize=4                   %
]
You are an expert in clinical documentation. You will be given a checklist question and a medical note that fulfills the criteria, with respect to the given clinical transcript. 
Your task is to rewrite the note so that it fails the criteria with respect to the transcript.
\end{lstlisting}

User message for re-writing reference notes to fail checklist criteria:
\begin{lstlisting}[
  basicstyle=\ttfamily\small, %
  breaklines=true,            %
  frame=single,               %
  columns=flexible,           %
  keepspaces=true,            %
  tabsize=4                   %
]
Transcript: {reference transcript}

Note: {reference note}

Question: {question}
\end{lstlisting}

\subsection{Feedback Coverage}
\label{assec:coverage}

Assigner's system message:
\begin{lstlisting}[
  basicstyle=\ttfamily\small, %
  breaklines=true,            %
  frame=single,               %
  columns=flexible,           %
  keepspaces=true,            %
  tabsize=4                   %
]
You will be given an itemized checklist of questions and a user feedback. 
Indicate which questions cover the feedback. Respond as a list of question item numbers.

Questions: 
{itemized checklist questions}
\end{lstlisting}

\begin{figure}
    \centering
    \includegraphics[width=0.95\linewidth]{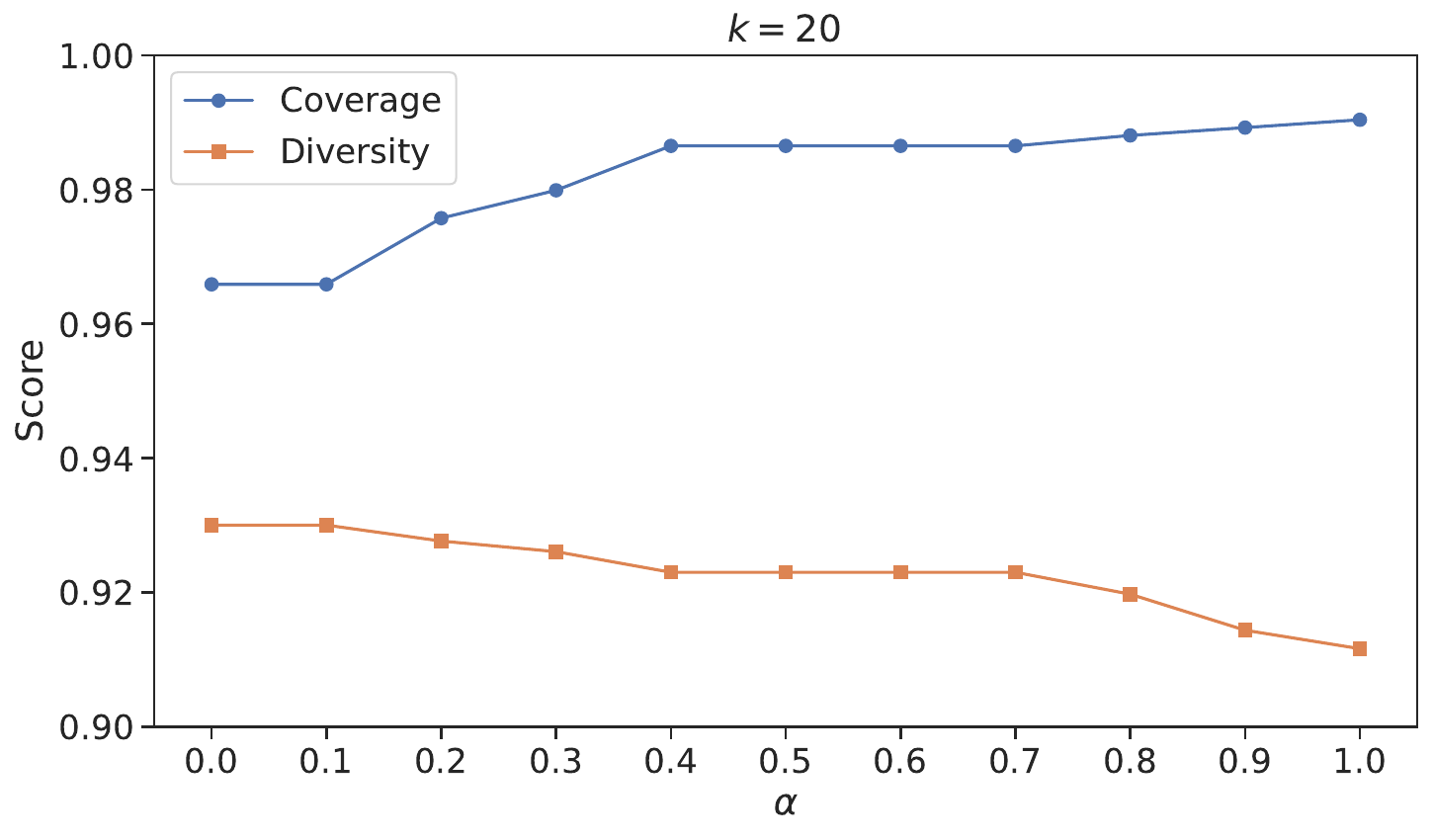}
    \caption{$\alpha$ values for the objective function, where $\alpha$ is the weight of the coverage term. The $\alpha$ value is set to $0.5$ for the final checklists, since it provides balance between coverage and diversity.\label{fig:alphas}}
\end{figure}

\section{Extended Refinement Details}\label{asec:refinement}

\subsection{Examples of Flagged Questions}\label{assec:flagged_questions}
When tagging for applicability, we drop questions like ``Is there a comprehensive assessment and plan for all chronic conditions discussed, ensuring nothing important is omitted?'' 
For section-specificity, we flag questions like ``Is there a clear separation between the history of present illness (HPI) and the action plan to ensure there is no overlap?'' since they would require access to the {\it HPI} section to answer.

\subsection{LLM Enforceability}\label{assec:enforce}
The 10 reference notes in each question's mini-benchmark are selected randomly among all the ones that pass the question. We prompt \verb|gpt-4.1-mini| to rewrite the reference notes to fail; see \secref{assec:ut} for the prompt used to rewrite the reference notes to fail the checklist questions. For example, if the checklist question is ``Does the note use the correct pronouns?'' and the reference note uses ``she'', then the note could be rewritten to use ``he'' to fail the criteria.

Post-processing is done to ensure only the rewritten note is produced, i.e., no explanations. Manual verification of 50 pairs of notes spanning 27 different checklist questions has yielded a 90\% validation rate (rewritten notes that correctly fail); the remaining 10\% of rewritten notes are still of inferior quality (missing/inaccurate details), but do not necessarily fail the criteria of the question.

\subsection{Optimizing for Feedback Coverage and Diversity}\label{assec:objective}

Ideally, the checklists questions should maximize coverage of user feedback while avoiding redundancy. 
This refinement step accounts for addressing the most salient feedback points, while avoiding questions that are too specific to a single encounter or too general and uninformative. 

\textit{Coverage} is defined as whether each feedback is supported by each question, and it is measured by the assigner LLM, with the prompt defined in \secref{assec:coverage}. We obtain a ``coverage matrix'', where $(i,j)$ value is 1 if question $i$ covers feedback $j$, and 0 otherwise. Total coverage rate $C$, is the sum of the coverage matrix divided by the number of feedback items.

To avoid redundancy, we also want feedback \textit{diversity} among the checklist questions' coverage sets. For questions $i$ and $j$ with covered sets of feedback $F_i$ and $F_j$, we define the \textit{similarity} $S_{ij}$ of the feedback items covered by questions $i$ and $j$ as their Jaccard similarity, i.e., $S_{ij} = |F_i \cap F_j| / |F_i \cup F_j|$.
Then for a checklist of $n$ questions, we can define the diversity of question $i$ as one minus its average similarity with all other questions:
\begin{equation*}
    D_i = 1 - \frac{1}{n-1} \sum_{j=1, j\ne i}^{n}  S_{ij}
\end{equation*}

Total diversity $D$ is the average of the diversity of all questions. 

Since we also want to avoid overly length checklists, we also add a weight $\lambda$ for penalizing length. 
Finally, for a checklist of $k$ questions, we can define a objective score to balance coverage and diversity of a checklist, given $\alpha$ as the weight:
\begin{equation*}
    Score(k) = \alpha \cdot C + (1-\alpha) \cdot D - \lambda \cdot k
\end{equation*}

We can then use beam search to select the optimal subset of $k$ questions.
\figref{fig:alphas} shows the $\alpha$ values for the objective function, where $\alpha$ is the weight of the coverage term. We select a final $\alpha$ value of $0.5$, since it provides balance between coverage and diversity.

\begin{figure}
    \centering
    \includegraphics[width=0.95\linewidth]{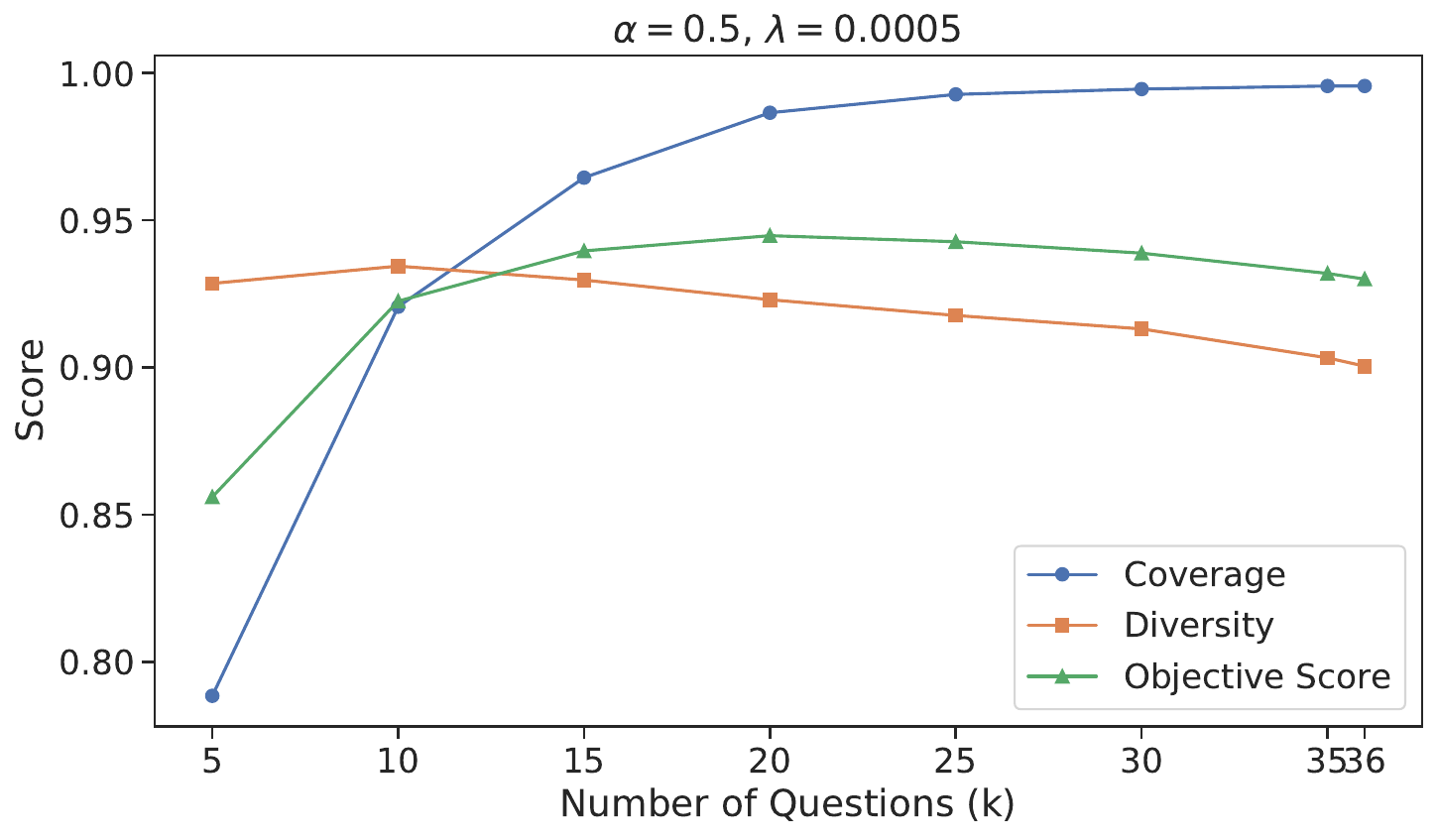}
    \caption{Objective score (coverage and diversity) vs. length of checklist. $k=20$ is the optimal number of questions.
    }
    \label{fig:coverage_v_k}
    \label{fig:feedback_coverage_diversity}
\end{figure}

\begin{figure*}
    \centering
   \includegraphics[width=0.85\linewidth]{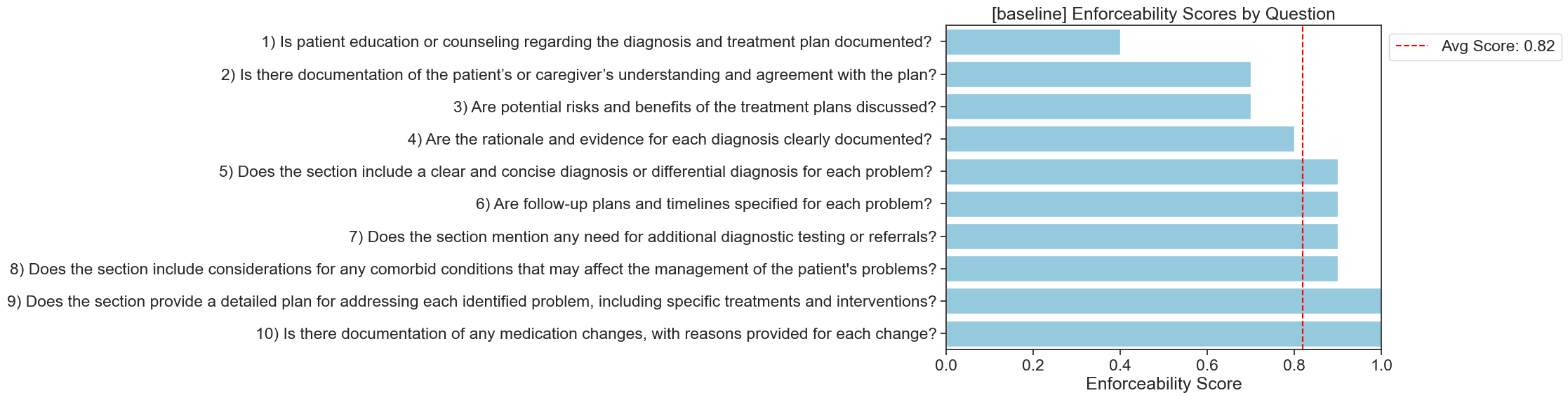}
    \includegraphics[width=0.95\linewidth]{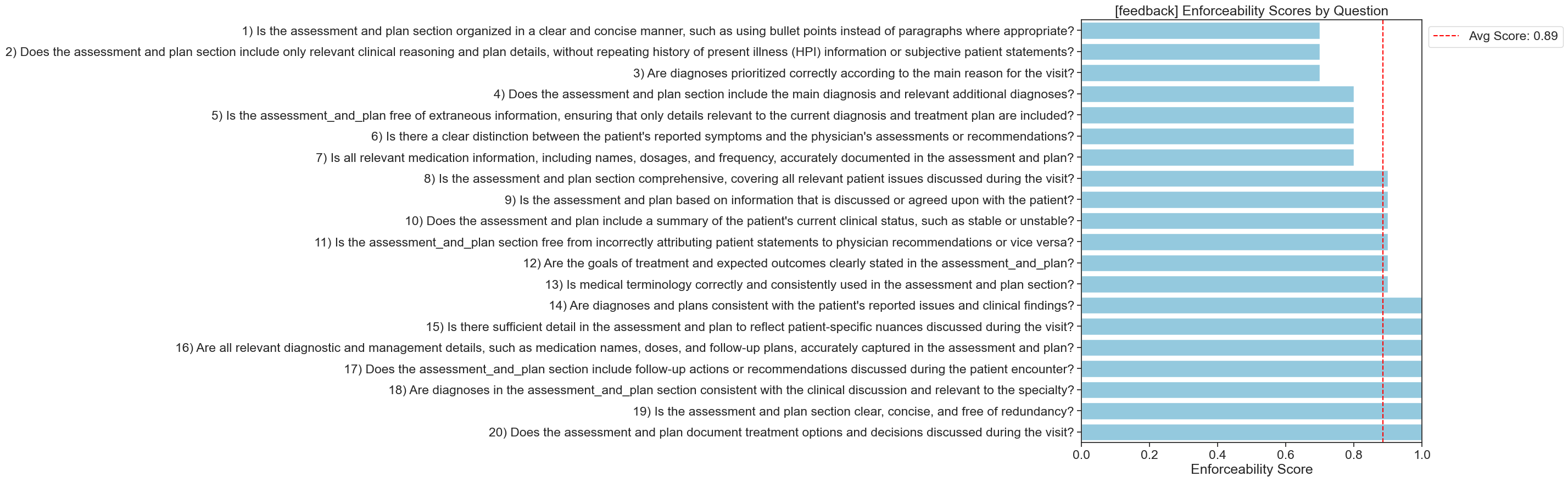}
    \caption{The enforceability score of each checklist question, indicating how LLM enforceable each one is. The feedback checklist has both a greater quantity of perfectly enforceable questions and a higher average rate of enforceability.}
    \label{fig:unit_test_scores}
\end{figure*}

 \begin{figure}
      \centering
    \includegraphics[width=0.9\linewidth]{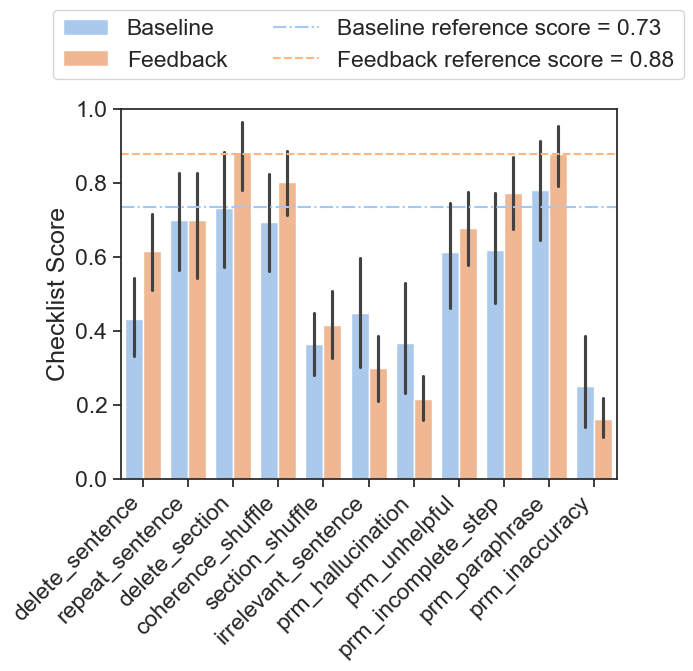}
    \caption{ The feedback checklist has a higher average score for the reference notes than the baseline checklist.}
    \label{fig:perturb_scores}
    \end{figure}

\begin{table*}[h!]
\centering
\footnotesize
\begin{tabular}{llll}
\toprule
\textbf{Name} & \textbf{Description} & \textbf{Section} & \textbf{\# Notes} \\
\cmidrule(lr){1-4}
\verb|delete_sentence|      & delete random sentences from each section             & All & 207 \\
\verb|delete_section|       & remove a whole section of the note                    & All & 207 \\
\verb|repeat_sentence|      & repeat random sentences in each section              & All & 207 \\
\verb|coherence_shuffle|    & shuffle sentences within each section                & All & 207 \\
\verb|section_shuffle|      & shuffle sentences across sections                    & All & 207 \\
\verb|irrelevant_sentence|  & insert irrelevant sentences (e.g., from other notes) & All & 207 \\
\verb|prm_inaccuracy|       & errors involving incorrect info or unsupported topics& AP  & 67  \\
\verb|prm_hallucination|    & introduce unrelated subject entities                 & AP  & 67  \\
\verb|prm_unhelpful|        & introduce vague, confusing, or incomplete expressions & AP  & 67  \\
\verb|prm_incomplete_step|  & randomly removing specific steps from A\&P            & AP  & 67  \\
\verb|prm_paraphrase|       & Gemini Pro 1.5 paraphrased note sections             & AP  & 67  \\
\bottomrule
\end{tabular}
\caption{Descriptions of the perturbed notes for evaluation. The number of notes is the number of reference notes that were perturbed with each perturbation. Perturbations beginning with \texttt{prm} are from \citet{wang2025processsupervisedrewardmodelsverifying}}
\label{tab:perturbations}
\end{table*}

\section{Checklist Questions} 
\label{asec:checklists}
The checklists are intended for quality assessment, not reimbursement or coding optimization; additional audit measures are required to prevent misuse.

\paragraph{Baseline Checklist for \textit{AP} ($n=10$)} 
\begin{enumerate}[leftmargin=*, itemsep=0em]
    \item Does the section include a clear and concise diagnosis or differential diagnosis for each problem?
    \item Are the rationale and evidence for each diagnosis clearly documented?
    \item Does the section provide a detailed plan for addressing each identified problem, including specific treatments and interventions?
    \item Are follow-up plans and timelines specified for each problem?
    \item Does the section mention any need for additional diagnostic testing or referrals?
    \item Is patient education or counseling regarding the diagnosis and treatment plan documented?
    \item Is there documentation of the patient's or caregiver's understanding and agreement with the plan
    \item Does the section include considerations for any comorbid conditions that may affect the management of the patient's problems?
    \item Is there documentation of any medication changes, with reasons provided for each change?
    \item  Are potential risks and benefits of the treatment plans discussed?
\end{enumerate}

\paragraph{Feedback Checklist for \textit{AP} ($n=20$)}

\begin{enumerate}[leftmargin=*, itemsep=0em]
  \item Is the assessment and plan section organized in a clear and concise manner, such as using bullet points instead of paragraphs where appropriate?
\item Is the assessment and plan section comprehensive, covering all relevant patient issues discussed during the visit?
\item Are diagnoses and plans consistent with the patient's reported issues and clinical findings?
\item Does the assessment and plan section include the main diagnosis and relevant additional diagnoses?
\item Is there sufficient detail in the assessment and plan to reflect patient-specific nuances discussed during the visit?
\item Does the assessment and plan section include only relevant clinical reasoning and plan details, without repeating history of present illness (HPI) information or subjective patient statements?
\item Are diagnoses prioritized correctly according to the main reason for the visit?
\item Is the assessment and plan based on information that is discussed or agreed upon with the patient?
\item Are all relevant diagnostic and management details, such as medication names, doses, and follow-up plans, accurately captured in the assessment and plan?
\item Does the assessment and plan include a summary of the patient's current clinical status, such as stable or unstable?
\item Is the assessment and plan section free from incorrectly attributing patient statements to physician recommendations or vice versa?
\item Does the assessment and plan section include follow-up actions or recommendations discussed during the patient encounter?
\item Is the assessment and plan free of extraneous information, ensuring that only details relevant to the current diagnosis and treatment plan are included?
\item Are diagnoses in the assessment and plan section consistent with the clinical discussion and relevant to the specialty?
\item Is there a clear distinction between the patient's reported symptoms and the physician's assessments or recommendations?
\item Are the goals of treatment and expected outcomes clearly stated in the assessment and plan?
\item Is the assessment and plan section clear, concise, and free of redundancy?
\item Does the assessment and plan document treatment options and decisions discussed during the visit?
\item Is all relevant medication information, including names, dosages, and frequency, accurately documented in the assessment and plan?
\item Is medical terminology correctly and consistently used in the assessment and plan section?
\end{enumerate}

\section{Additional Evaluation and Results}
\label{asec:eval_results}

\subsection{Generation and Refinement}
\label{assec:gen_refine}

The \textasciitilde$7700$ 
feedback items for the {\it AP} section are randomly split into 8 batches of approximately 900-1000 items each. 97 candidate checklist questions are generated from this set of feedback.

Filtering out redundant questions and tag for applicability and specificity results in 50 questions.
For example, the questions 
    ``Is the assessment and plan section well-organized and clearly separates different medical issues or diagnoses?'' and 
    ``Is the assessment and plan organized in a way that is easy to read, with distinct separation between different patient problems?'' are clustered and merged into a single question: ``Is the assessment and plan section well-organized with clear separation of medical issues or diagnoses?''

These 50 questions are then evaluated for LLM enforceability. We find that 36 questions have a unit test threshold of at least 0.7, and we discard the remaining 14 questions. Such discarded questions include: ``Is unnecessary and speculative language avoided in the assessment and plan section?'' The average enforceability score for the remaining questions is 0.9.

Finally, we select a subset of $k$ questions that maximize coverage and diversity. Setting $\alpha=0.5$ and $\lambda=0.0005$, we find that $k=20$ provides the optimal balance of coverage and diversity with length of the checklist (see \figref{fig:coverage_v_k}).

\subsection{Evaluation}
\label{assec:eval_metrics}

\paragraph{Predictive Power}
For simplicity, we set up the task as a binary classification task, by using data with star ratings of 1 as negative and 5 as positive (1113 and 680 samples respectively). 
We then train a Logistic Regression model with 5-fold cross-validation on 80\% of the feedback and test with the remaining 20\%. 

\paragraph{LLM Enforceability Results}
 From \figref{fig:unit_test_scores}, we see that the baseline checklist has an average unit test scores of 0.82 over 10 questions, while the feedback checklist has an average unit test score of 0.89 over 20 questions.

\paragraph{Robustness to Perturbations}

We are inspired by the perturbations defined by \citet{guo2024applsevaluatingevaluationmetrics} for our use case, though we implement our own methods. We also use data from \citet{wang2025processsupervisedrewardmodelsverifying} as perturbed versions of {\it AP} note sections.
Specifically, we apply each of the following perturbations in \tbref{tab:perturbations} to the reference notes.
With this benchmark, we can measure the robustness of our final checklists against changes in informativeness, coherence, and factuality.

We also see that our feedback checklist produces a higher average score for the reference notes than the baseline checklist (\figref{fig:perturb_scores}).

\section{Future Work}
\label{asec:future}
Some evaluation metrics, like feedback coverage/diversity and LLM enforceablity, can be incorporated into the checklist refinement pipeline, to help select for the most salient questions. This would allow the refinement process to be more iterative.

\paragraph{Scaling the Pipeline}
\label{ssec:scaling}
We would like to scale the pipeline to generate checklists for other note sections.
We would also like to implement a dynamic pipeline, where the checklist is updated as new feedback comes in. This would allow us to continuously improve the checklist and adapt it to changing user needs.

\paragraph{Additional Feedback Filtering}
The feedback we receive is often unstructured and noisy. There is a lot of redundancy in the feedback, and not all of it is actionable. So, we can apply additional filtering steps to the feedback, such as grouping by specialty, filtering out un-actionable feedback, and clustering similar feedback.

\paragraph{Question Importance for Predicting Human Ratings} We can further identify questions that have the greatest influence on performance, using feature importance measures like SHAP values \citep{NIPS2017_7062}. These values can also help assign priority to the order of questions in the checklist, so that the most important questions are asked first or given more weight in the final checklist score.

\paragraph{Human Evaluation}
To validate the quality of our checklists, we will run human evaluation. Some potential user studies include:

\begin{itemize}
    \item Pairwise preference of checklists (e.g., rate baseline vs. feedback checklist questions)
    \item Have physicians conduct evaluation of real deployed notes 
    using checklists, and see if there is higher agreement between raters compared to evaluationg without a checklist
\end{itemize}

\paragraph{Generalizability}
To show the generalizability of our checklist generation method, we would like to apply it to some alternate domain. We are still brainstorming what dataset(s) we could use for this. Requirements of the data include user feedback and ratings for some product that should fulfill certain standards.

\paragraph{Tuning a LLM-evaluator}
We can further try tuning LLM-as-a-Judge specifically for our task of checklist scoring. While the current setup simply has the LLM-evaluator produce a ``Yes''/``No'' answer for each checklist question, we can also allow the LLM to reason about the note and checklist question and provide a more detailed explanation of its score. 
Allowing the evaluator to cross-validate the note with structured EHR data could enhance accuracy by catching factual inconsistencies that text-only evaluation misses.

\end{document}